\documentclass[review]{elsarticle}
\usepackage{geometry}
\usepackage{lineno,hyperref}
\usepackage{booktabs}
\usepackage{graphicx}
\usepackage{multicol}
\usepackage{adjustbox}
\usepackage{multirow}
\usepackage{caption}
\usepackage{amsmath}
\usepackage{subcaption}
\usepackage{color}
\usepackage{fancyhdr}
\usepackage{array}
\usepackage[flushleft]{threeparttable}
\usepackage{rotating}
\usepackage[withpage]{acronym}
\modulolinenumbers[5]

\journal{Journal}

\biboptions{authoryear}

\begin{document}

\begin{frontmatter}

\title{Deep Learning in Multimodal Remote Sensing Data Fusion: A Comprehensive Review}

\author[firstaddress,thirdaddress]{Jiaxin Li}
\ead{lijiaxin203@mails.ucas.ac.cn}

\author[firstaddress]{Danfeng Hong}
\ead{hongdf@aircas.ac.cn}

\author[firstaddress]{Lianru Gao\corref{correspondingauthor}}
\cortext[correspondingauthor]{Corresponding author}
\ead{gaolr@aircas.ac.cn}

\author[firstaddress]{Jing Yao}
\ead{yaojing@aircas.ac.cn}

\author[fouraddress,firstaddress]{Ke Zheng}
\ead{zhengkevic@aircas.ac.cn}

\author[secondaddress,thirdaddress]{Bing Zhang}
\ead{zb@radi.ac.cn}

\author[fifthaddress,secondaddress]{Jocelyn Chanussot}
\ead{jocelyn@hi.is}

\address[firstaddress]{Key Laboratory of Computational Optical Imaging Technology, Aerospace Information Research Institute, Chinese Academy of Sciences, Beijing 100094, China;}
\address[secondaddress]{Aerospace Information Research Institute, Chinese Academy of Sciences, Beijing 100094, China;}
\address[thirdaddress]{College of Resources and Environment, University of Chinese Academy of Sciences, Beijing 100049, China;}
\address[fouraddress]{College of Geography and Environment, Liaocheng University, Liaocheng, 252059, China;}
\address[fifthaddress]{University Grenoble Alpes, CNRS, Grenoble INP, GIPSA-Lab,  Grenoble 38000, France.}

\begin{abstract}
With the extremely rapid advances in remote sensing (RS) technology, a great quantity of Earth observation (EO) data featuring considerable and complicated heterogeneity is readily available nowadays, which renders researchers an opportunity to tackle current geoscience applications in a fresh way. With the joint utilization of EO data, much research on multimodal RS data fusion has made tremendous progress in recent years, yet these developed traditional algorithms inevitably meet the performance bottleneck due to the lack of the ability to comprehensively analyse and interpret these strongly heterogeneous data. Hence, this non-negligible limitation further arouses an intense demand for an alternative tool with powerful processing competence. Deep learning (DL), as a cutting-edge technology, has witnessed remarkable breakthroughs in numerous computer vision tasks owing to its impressive ability in data representation and reconstruction. Naturally, it has been successfully applied to the field of multimodal RS data fusion, yielding great improvement compared with traditional methods. This survey aims to present a systematic overview in DL-based multimodal RS data fusion. More specifically, some essential knowledge about this topic is first given. Subsequently, a literature survey is conducted to analyse the trends of this field. Some prevalent sub-fields in the multimodal RS data fusion are then reviewed in terms of the to-be-fused data modalities, i.e., spatiospectral, spatiotemporal, light detection and ranging-optical, synthetic aperture radar-optical, and RS-Geospatial Big Data fusion. Furthermore, We collect and summarize some valuable resources for the sake of the development in multimodal RS data fusion. Finally, the remaining challenges and potential future directions are highlighted.

\end{abstract}

\end{frontmatter}


\begin{table}[!t]
\centering
\caption{List of the main abbreviations}
\label{tab:abbreviations}
\begin{adjustbox}{width=1\textwidth}
\begin{tabular}{llll}
\hline
Abbreviation & Description                    & Abbreviation & Description                            \\ \hline
AE           & Autoencoder                    & LULC         & Land use and land cover                \\ \hline
CS           & Component substitution         & LiDAR        & Light detection and ranging            \\ \hline
CNN          & Convolutional neural network   & MF           & Matrix factorization                   \\ \hline
DHP          & Deep hyperspectral prior       & MRA          & Multiresolution analysis               \\ \hline
DL           & Deep learning                  & MS           & Multispectral                          \\ \hline
DI           & Details injection              & NDVI         & Normalized difference vegetation index \\ \hline
EO           & Earth observation              & Pan          & Panchromatic                           \\ \hline
EP           & Extinction profile             & POI          & Points of interest                     \\ \hline
GAN          & Generative adversarial network & RS           & Remote sensing                         \\ \hline
GBD          & Geospatial big data            & SAR          & Synthetic aperture radar               \\ \hline
GNN          & Graph neural network           & TR           & Tensor representation                  \\ \hline
HS           & Hyperspectral                  & VO           & Variational optimization               \\ \hline
LST          & Land surface temperature       & ViT          & Visual transformer                     \\ \hline
\end{tabular}
\end{adjustbox}
\end{table}

\section{Introduction} \label{Introduction}

\journal{}
On account of the superiority in observing our Earth environment, RS has been playing an increasingly important role in various EO tasks \citep{hong2021interpretable, zhang2019remotely}. With the ever-growing availability of multimodal RS data, researchers have easy access to the data, which is suitable for the application at hand. Although a large amount of multimodal data become readily available, each modality can barely capture one or few specific properties and hence cannot fully describe the observed scenes, which poses a great constraint on subsequent applications. Naturally, multimodal RS data fusion is a feasible way to break out of the dilemma induced by unimodal data. By integrating the complementary information extracted from multimodal data, a more robust and reliable decision can be made in many tasks, such as change detection, LULC classification, etc.

Unlike multisource and multitemporal RS, the term of  ``modality'' has been a lack of a clear and unified definition. 
In this paper, we attempted to give a detailed definition on basis of previous works \citep{gomez2015multimodal, dalla2015challenges}. Principally, RS data is characterized by two main factors, i.e., the technical specifications of the sensors and the actual acquisition condition. Specifically, the former determine the internal characteristics of the product, e.g., imaging mechanism and the resolutions in the domain of spatial, spectral, radiometric, and temporal. While, the latter control the external properties, e.g., the acquisition time, observation angles, and mounted platforms. Thus, the aforementioned factors contribute to the descriptions of the captured scene and can be described as ``modality''. Apparently, multimodal RS data fusion include multisource data fusion and multitemporal data fusion.

\begin{figure}[!t]
\centering
\includegraphics[width=\linewidth]{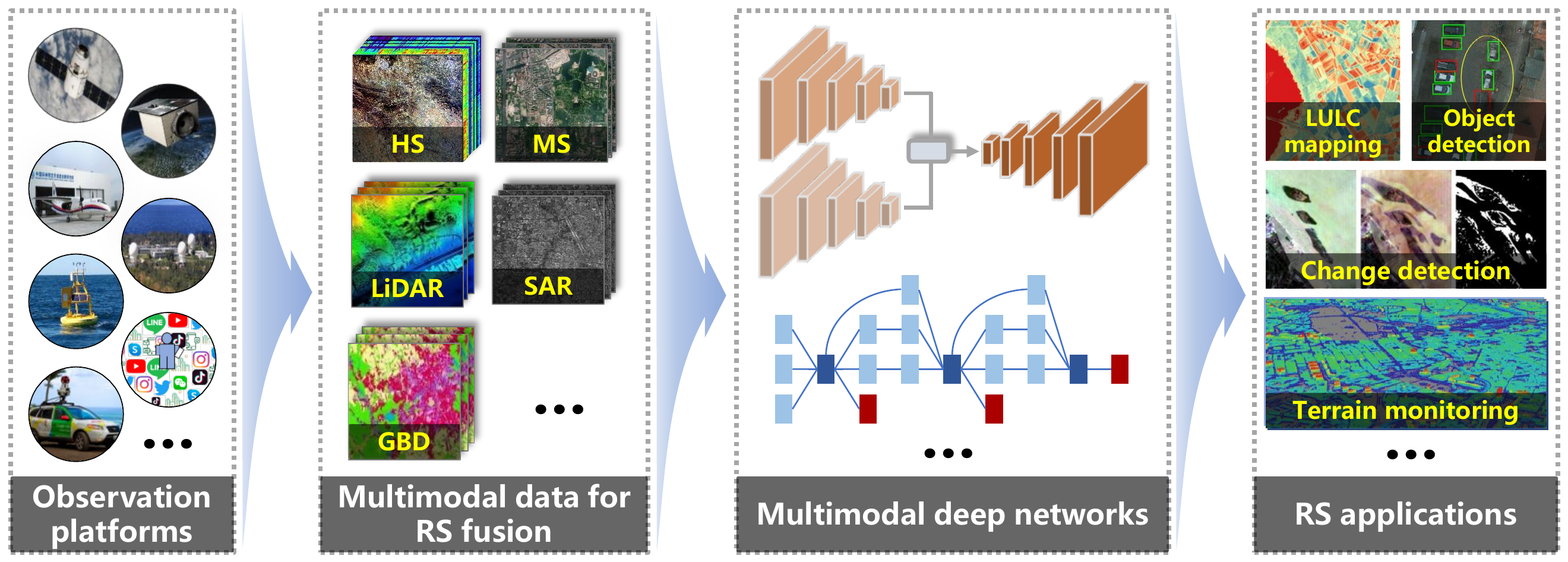}
\caption{An illustration of DL in multimodal RS data fusion.}
\label{fig:framework}
\end{figure}

Some typical RS modalities include Pan, MS, HS, LiDAR, SAR, infrared, night time light, and satellite video data. Very recently, GBD, as a new member in the RS family, has attracted growing attention in the EO tasks. To integrate the complementary information provided by these modalities, traditional methods have been intensively studied by designing handcrafted features based on domain-specific knowledge and exploiting rough fusion strategies, which inevitably impairs the fusion performance, especially for heterogeneous data \citep{hong2021more}. Thanks to the growth of artificial intelligence, DL shows great potential in modelling a complicated relationship between input and output data by adaptively realizing the feature extraction and fusion in an automatic manner. 
Depending on the to-be-fused modalities and corresponding tasks, DL-based multimodal RS data fusion can be generalized into a unified framework (see figure \ref{fig:framework}). Accordingly, this review will focus on the methods proposed in each fusion subdomain along with a brief introduction in each modality and related tasks.

Currently, there exist some literature reviews regarding multimodal data fusion, which are summarized in table \ref{tab:classical review} according to different modality fusion. Existing reviews either pay less attention to the direction of DL or only cover few sub-areas in multimodal RS data fusion, lacking a comprehensive and systematic description on this topic. The motivation of our survey is to give a comprehensive review of popular domains in DL-based multimodal RS data fusion, and further facilitate and promote the relevant research in this burgeoning domain. More specifically, literature related to this topic is collected and analyzed in section \ref{statistical}, followed by section \ref{methods}, which elaborates on representative sub-fields in multimodal RS data fusion. In section \ref{List of resources}, some useful resources in respect of tutorials, datasets and codes are given. Finally, section \ref{Problems and prospects} provides remarks concerning the challenges and prospects. For the convenience of readers, main abbreviations used in this article is listed in table \ref{tab:abbreviations}.

\begin{table}[!t]
\centering
\caption{Typical multimodal data fusion reviews}
\label{tab:classical review}
\begin{adjustbox}{width=1\textwidth}
\begin{tabular}{clll}

\hline
\multicolumn{1}{l}{}    & Domains                                 & References & Descriptions                                                                                                          \\ \hline
\multirow{11}{*}{Homogeneous fusion}  & \multirow{4}{*}{Pansharpening}          & \citep{ranchin2003image}       & Introducing the methods belonging to ARSIS, along with giving a simple comparison                                     \\ \cline{3-4} 
                                      &                                         & \citep{vivone2014critical}       & Giving a thorough descripitions and assessments of the methods belonging to CS and MRA families                       \\ \cline{3-4} 
                                      &                                         & \citep{meng2019review}       & Introducing the methods belonging to CS, MAR, and VO from the idea of meta-analysis                                   \\ \cline{3-4} 
                                      &                                         & \citep{vivone2020new}       & Giving a systematic introduction and evaluation of the methods in the category of CS, MAR, VO, and ML                 \\ \cline{2-4} 
                                      & HS pansharpening                        & \citep{loncan2015hyperspectral}       & Conducting a comprehensive analysis and evaluation in the methods from CS, MAR, hybrid, bayesian, and MF              \\ \cline{2-4} 
                                      & \multirow{2}{*}{HS-MS fusion}           & \citep{yokoya2017hyperspectral}       & Extensive experiments are presented to assess the methods from CS, MRA, unmixing, and bayesian                        \\ \cline{3-4} 
                                      &                                         & \citep{dian2021recent}       & Studying the performance of methods from CS, MAR, MF, TR, and DL                                                      \\ \cline{2-4} 
                                      & \multirow{4}{*}{Spatiotemporal}         & \citep{chen2015comparison}       & Discussing and evaluating four models from transformation/reconstruction/learning-based methods                       \\ \cline{3-4} 
                                      &                                         & \citep{zhu2018spatiotemporal}       & Reviewing the characteristics of five categories and their applications                                               \\ \cline{3-4} 
                                      &                                         & \citep{belgiu2019spatiotemporal}       & Introducing the methods in three categories, as well as the challenges and opportunities                              \\ \cline{3-4} 
                                      &                                         & \citep{li2020spatio}       & Analyzing the performance of representative methods with their provided benchmark dataset                             \\ \hline
\multirow{5}{*}{Heterogeneous fusion} & \multirow{2}{*}{HS-LiDAR}               & \citep{man2014light}       & Summarizing the research on HS-LiDAR fusion for forest biomass estimation                                             \\ \cline{3-4} 
                                      &                                         & \citep{kuras2021hyperspectral}       & Giving an overview of HS-LiDAR fusion in the application of land cover classification                                 \\ \cline{2-4} 
                                      & SAR-optical                             & \citep{kulkarni2020pixel}       & Evaluating the performance of methods in CS and MRA in pixel-level                                                    \\ \cline{2-4} 
                                      & \multirow{2}{*}{RS-GBD} & \citep{li2021distributed}       & Providing a review on RS-social media fusion and their distributed strategies                                         \\ \cline{3-4} 
                                      &                                         & \citep{yin2021integrating}       & Reviewing the fusion of RS-GBD in the application of urban land use mapping from feature-level and decision-level perspectives\\ \hline
\multicolumn{2}{c}{\multirow{11}{*}{Others}}                         & \citep{wald1999some}       & Setting up some definitions regrading data fusion                                                                     \\ \cline{3-4} 
\multicolumn{2}{c}{}                                                            & \citep{gomez2015multimodal}       & Providing a review in seven data fusion applications for RS                                                           \\ \cline{3-4} 
\multicolumn{2}{c}{}                                                            & \citep{lahat2015multimodal}       & Summarizing the challenges in multimodal data fusion across various disciplines                                       \\ \cline{3-4} 
\multicolumn{2}{c}{}                                                            & \citep{dalla2015challenges}       & Giving a comprehensive discussion on data fusion problems in RS by analyzing the Data Fusion Contests                                                     \\ \cline{3-4} 
\multicolumn{2}{c}{}                                                            & \citep{ghassemian2016review}      & Introducing the RS fusion methods in pixel/feature/decision-level and different evaluation criteria                   \\ \cline{3-4} 
\multicolumn{2}{c}{}                                                            & \citep{schmitt2016data}       & Modeling the data fusion process, along with introducing some typical fusion scenarios in RS                          \\ \cline{3-4} 
\multicolumn{2}{c}{}                                                            & \citep{li2017pixel}       & Introducing fusion methods in pixel-level and their major applications                                                \\ \cline{3-4} 
\multicolumn{2}{c}{}                                                            & \citep{liu2018deep}       & Reviewing DL-based pixel-level fusion methods in digital photography, multi-modality imaging, and RS imagery          \\ \cline{3-4} 
\multicolumn{2}{c}{}                                                            & \citep{ghamisi2019multisource}       & Conducting a detailed review in spatiospectral, spatiotemporal, HS-LiDAR, etc                                         \\ \cline{3-4} 
\multicolumn{2}{c}{}                                                            & \citep{zhang2021image}       & Reviewing DL-based fusion methods in digital photography, multi-modal image, sharpening fusion                        \\ \cline{3-4} 
\multicolumn{2}{c}{}                                                            & \citep{kahraman2021comprehensive}       & Describing methods in HS-LiDAR and HS-SAR fusion                                                                      \\ \hline
\end{tabular}
\end{adjustbox}
\end{table}


\section{Literature analysis}\label{statistical}

\subsection{Data retrieval and collection}
In this section, Web of Science and CiteSpace \citep{chen2006citespace} are chosen as the main analysis tools. Taking the Query one in Table \ref{tab:Data retrieval} for example, 691 results are initially returned from Web of Science Core Collection by using the advanced search: TS=(“remote sensing”) AND TS=(“deep learning”) AND TS=(“fusion”). After only considering the "Article" document type, 598 papers published from 2015 and to 2022 are included for the subsequent analysis.

\renewcommand\arraystretch{1.1}
\begin{table}[!htbp]
\centering
\centering
\caption{Data retrieval results of WOS from 2015 to 2022.}
\label{tab:Data retrieval}
\begin{adjustbox}{width=0.9\textwidth}
\begin{tabular}{l l l l}
\midrule

\multicolumn{1}{l}{Query}
&\multicolumn{1}{l}{Contents}
&\multicolumn{1}{l}{Original results}
&\multicolumn{1}{l}{Refined results}

\\
\hline
Q1 & (TS=(“remote sensing”) AND TS=(“deep learning”) AND TS=(“fusion”)) & 691 & 598  \\
Q2 & (TS=(“remote sensing”) AND TS=(“fusion”)) & 6483 & 4403         \\
\hline
\end{tabular}
\end{adjustbox}
\end{table}

\subsection{Statistical analysis and results}

\subsubsection{Statistical analysis of articles published annually}

\begin{figure}[!htbp]
\centering
\includegraphics[width=0.5\linewidth]{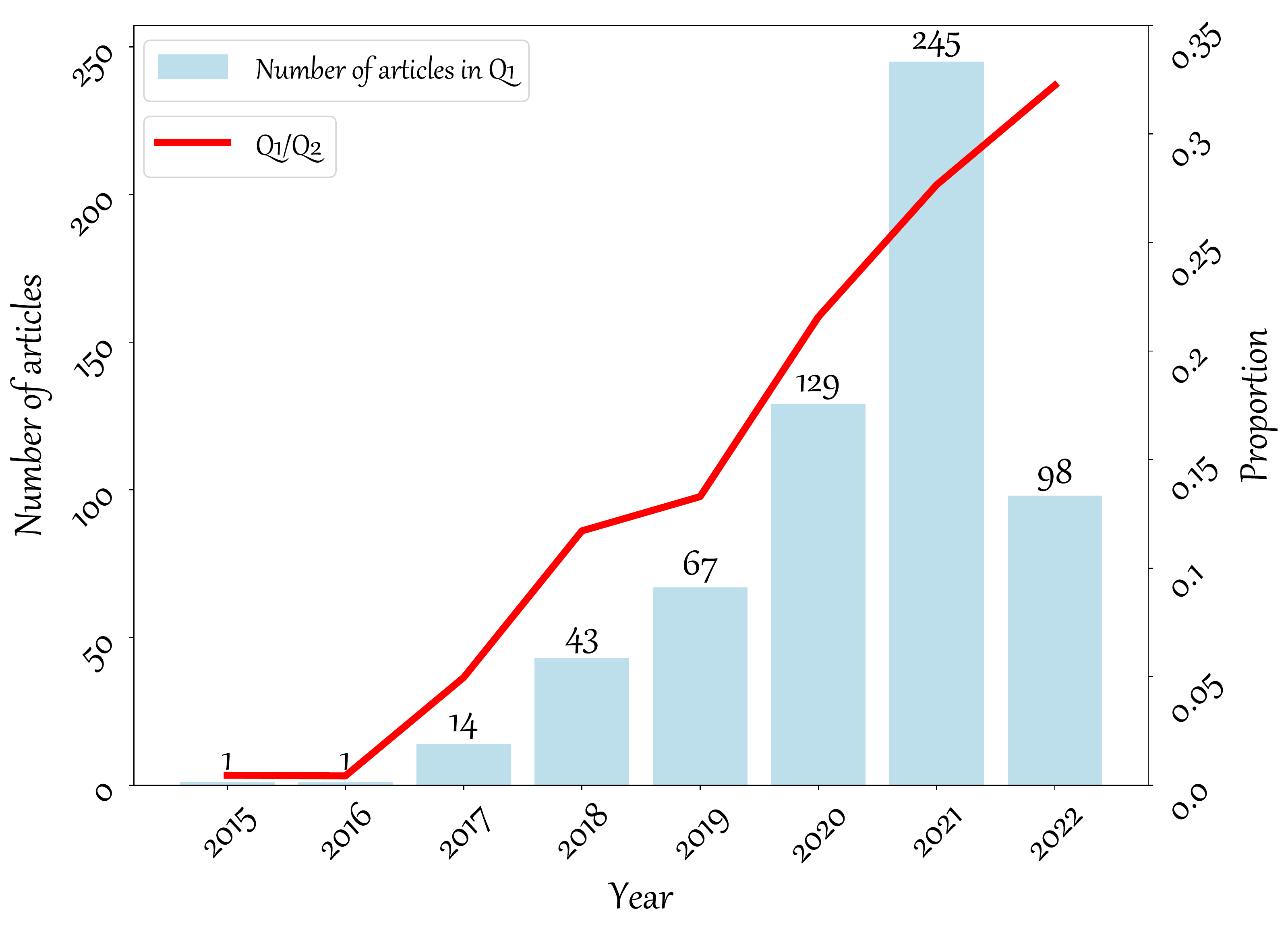}
\caption{Number of published articles annually on Q1 and its proportion on Q2.}
\label{fig:paper_annual}
\end{figure}

The trend of related papers published in 2015-2022 is shown in figure \ref{fig:paper_annual}. The bar chart suggests that growing attention has been paid to this burgeoning field with a steady increase in the number of publications. On the other hand, the upward trend in the line graph is consistent with that in the bar chart, which indicates that DL technologies have been playing an increasingly important role in the field of multimodal RS data fusion.

\subsubsection{Statistical analysis of the distribution of publications in terms of countries and journals}

\begin{figure}[!htbp]
\centering
\includegraphics[width=1\linewidth]{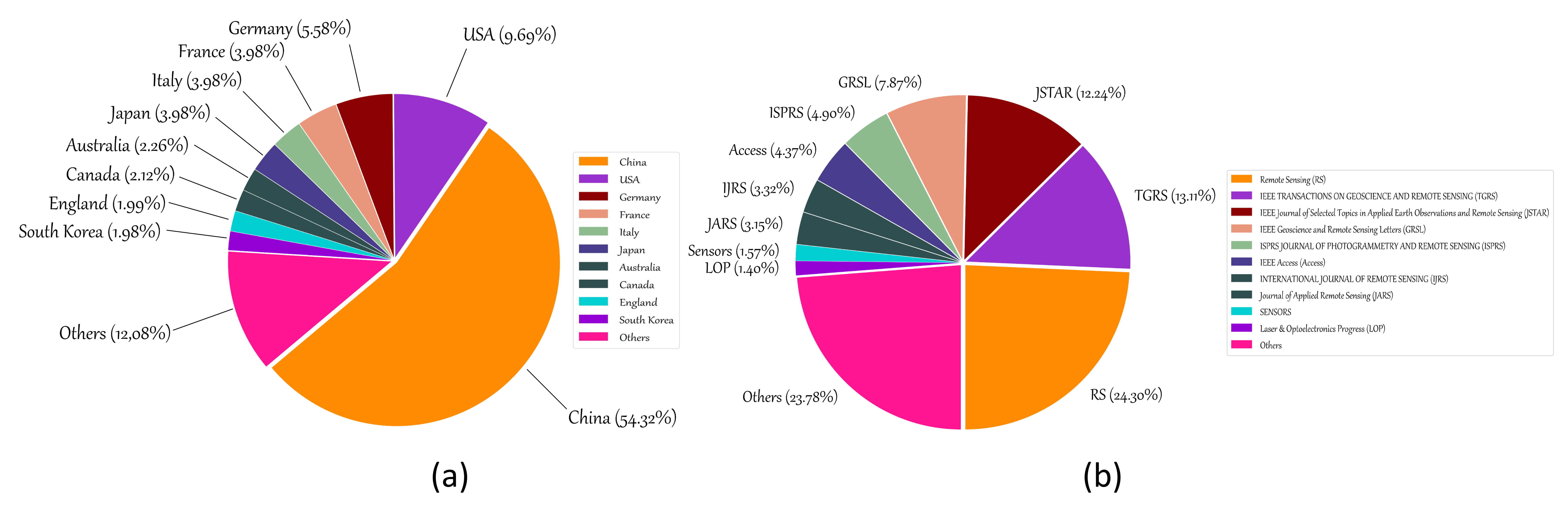}
\caption{Proportion of published articles by top 10 (a) countries and (b) journals.}
\label{fig:country_pie}
\end{figure}

Two pie charts showing the proportion of published papers by the top 10 countries and journals are displayed in figure \ref{fig:country_pie}(a) and figure \ref{fig:country_pie}(b), respectively. It can be seen that the top 10 countries take up about 90$\%$ of the total outputs, constituting the main pillar of this direction. More concretely, China makes a major contribution to the field, which accounts for more than half of all publications, followed by USA, which occupies about 10$\%$.

Besides, \emph{Remote Sensing}, \emph{IEEE TRANSACTIONS ON GEOSCIENCE AND REMOTE SENSING}, and \emph{IEEE Journal of Selected Topics in Applied Earth Observations and Remote Sensing} make up about half of the overall publications, with \emph{Remote Sensing} ranking first.




\subsubsection{Statistical analysis of the keywords in the literature}

\begin{figure}[!htbp]
\centering
\includegraphics[width=0.80\textwidth]{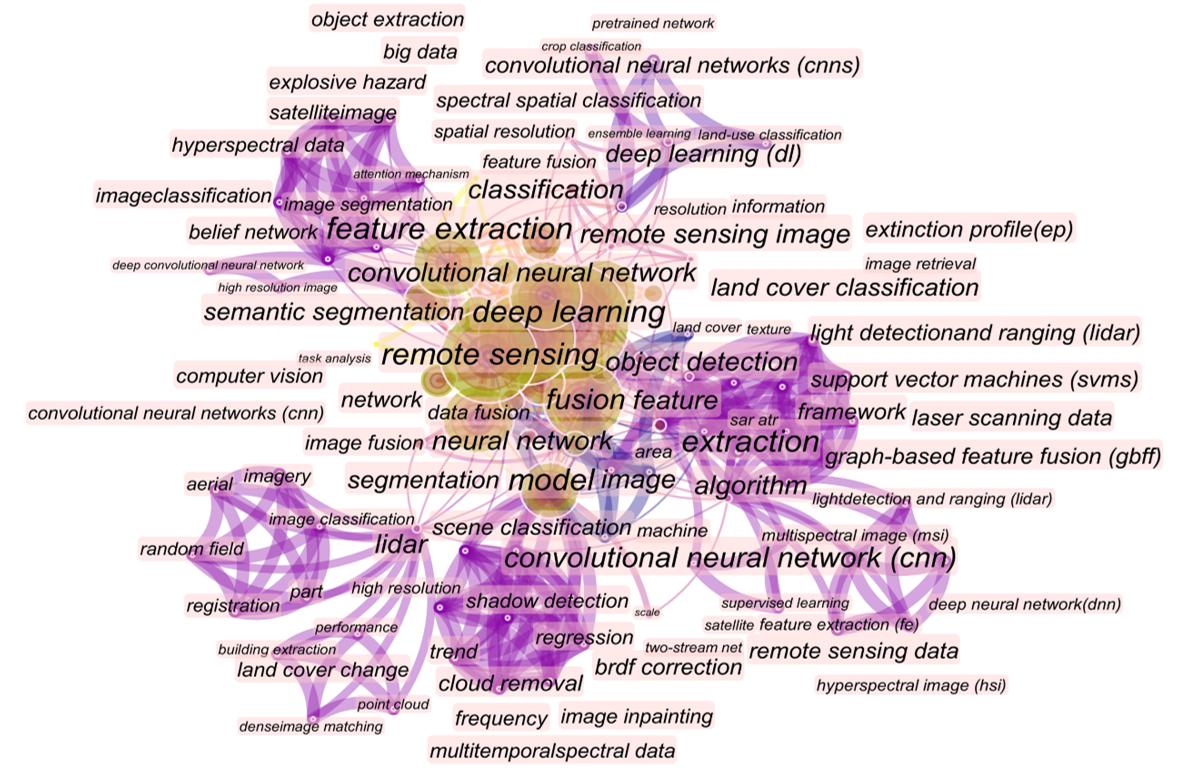}
\caption{A visualization of the keyword co-occurrence network.}
\label{fig:keyword}
\end{figure}

Figure \ref{fig:keyword} exhibits the keywords appearing in the collected articles, where a bigger font size corresponds to a higher frequency. As the figure indicates, CNN is widely used in the filed of DL-based multimodal RS data fusion. Besides, classification, cloud removal, and object detection become the main tasks in the fusion process, where MS, HS, LiDAR and SAR are the mainly-used data.

\section{A review of DL-based multimodal remote sensing data fusion methods}\label{methods}

\begin{figure}[!htbp]
\centering
\includegraphics[width=1\textwidth]{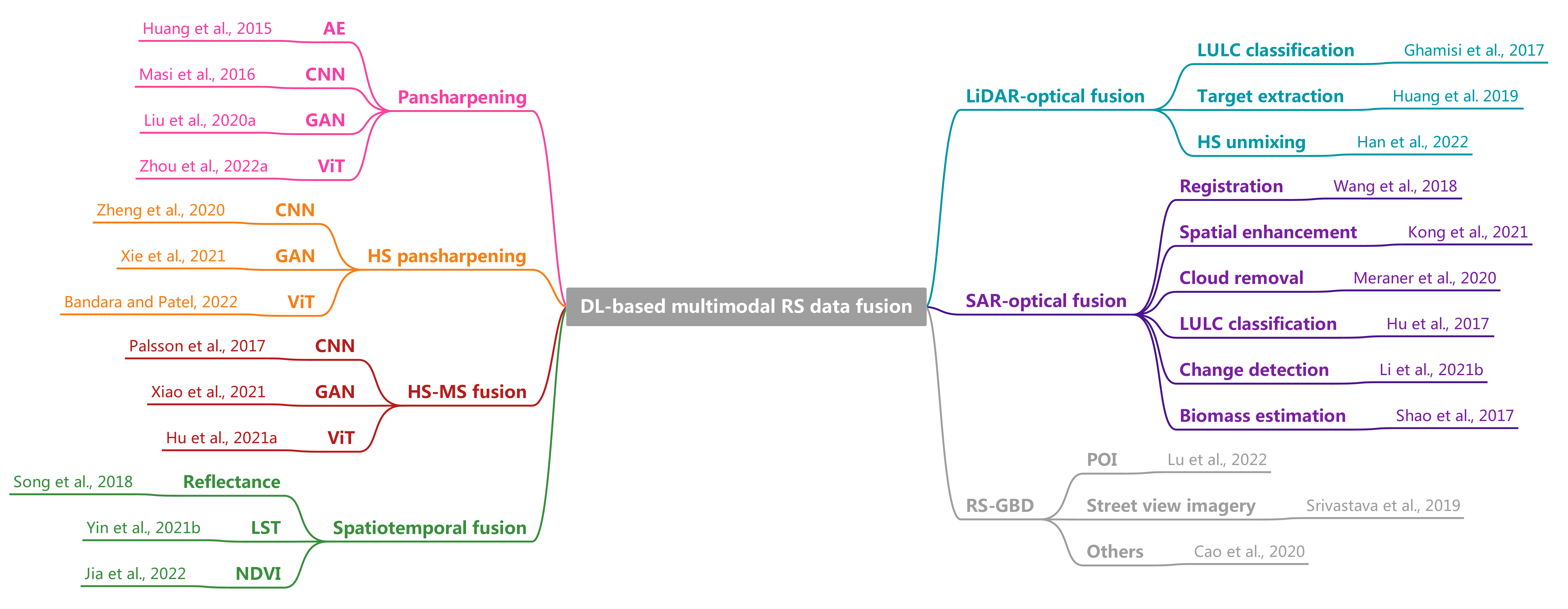}
\caption{The taxonomy of DL-based multimodal RS data fusion in this paper.}
\label{fig:taxonomy}
\end{figure}

This paper divide existing methods into two main groups, i.e., homogeneous fusion and heterogeneous fusion. Specifically, homogeneous fusion refers to pansharpening, HS pansharpening, HS-MS, and spatiotemporal fusion, while heterogeneous fusion includes HS-optical, SAR-optical, and RS-GBD fusion. Since the aforementioned sub-fields develop quite diversely, different criteria are adopted to introduce each subdomain, as shown in figure \ref{fig:taxonomy}. For the convenience of readers, we also list some classic literature in each direction.

\subsection{Homogeneous fusion}

The homogeneous fusion, including spatiospectral fusion (i.e., pansharpening, HS pansharpening, and HS-MS fusion) and spatiotemporal fusion, is primarily committed to solving the trade-off in spatial-spectral and spatial-temporal resolutions happening in the optical images due to the imaging mechanism. This section will introduce typical methods proposed in these domains.

\subsubsection{Pansharpening} \label{Pansharpening}

Pansharpening refers to the fusion of MS and Pan to generate a high spatial resolution Pan image. In general, AE, CNN, and GAN are commonly-used network architectures for DL-based pansharpening.

\begin{itemize}
\item \textbf{Supervised methods} 
\end{itemize}

It is well-known that supervised methods perform the pansharpening by linking the observations with the references. Usually, the input data need to be simulated by spatially downsampling the original data.
\cite{huang2015new} propose the first DL-based method in dealing with pansharpening problem, where a sparse denoising AE is adopted to learn the transformation in Pan domain, and then the observed MS is input into the pretrained AE to generate final output. Following this milestone work, many methods are successively proposed by treating pansharpening as an image super-resolution problem \citep{azarang2017new,xing2018pan}.
Apart from AE structure, CNN is also extensively used and can be categorized into three major groups, i.e., single-branch, multi-branch, and hybrid network. Methods belonging to the first group simply concatenate input Pan and up-sampled MS or their pre-processed versions into a new component as the input of network. For example, \cite{masi2016pansharpening} propose the first CNN-based pansharpening methods with three convolutional layers by adapting the SRCNN architecture. Later, numerous methods inspired by this pioneer work are presented, in which residual learning and dense connection are commonly used \citep{wei2017boosting,yang2017pannet,scarpa2018target,yuan2018multiscale,peng2020psmd,fu2020deep,lei2021nlrnet}. 
However, simply stacking pre-interpolated MS with Pan as the input of network not only ignores individual features but also raises extra computational burden. Hence, instead of treating the two modalities equally, multi-branch networks apply different sub-networks to separately extract the modality-specific features \citep{shao2018remote,zhang2019pan,liu2020remote,chen2021pansharpening,zhang2021gtp,xing2020dual,9503110}.
Hybrid network-based methods provide a cutting-edge solution to pansharpening by embracing the conception of traditional methods, i.e., DI-based methods \citep{he2019pansharpening,deng2020detail} and VO-based methods \citep{shen2019spatial,cao2021pancsc,tian2021vp}, and therefore effectively merging the strengths in both domains.
Different from CNN, GAN-based methods treat pansharpening as an image generation problem by establishing a adversarial game between a generator and a discriminator network. The first GAN-based pansharpening method designs a two-branch generator network \citep{liu2020psgan}, and then different loss functions and new network structures are explored to extract more discriminative features \citep{shao2019residual, ozcelik2020rethinking, gastineau2021generative}. 
ViT is recently introduced into pansharpening due to its ability in capturing long-range information \citep{zhou2021effective,zhou2022panformer}.  

\begin{itemize}
\item \textbf{Unsupervised methods} 
\end{itemize}

Scale-related problems may occur in supervised methods since they are often trained at lower resolution. However, unsupervised methods can implement the training and testing processes using original scale without need to simulate the references. Hence, the key of unsupervised methods is to precisely establish the relationships between the input data and fused product by designing proper loss functions, i.e., the degraded fusion result should be identical to input Pan and MS in spatial and spectral domains respectively. For example, \cite{ma2020pan} utilize a discriminator to preserve the spatial information using a gradient regularization between the input Pan and spectrally degraded version of the output from generator. The effective loss functions include gradient loss \citep{seo2020upsnet}, perceptual loss \citep{zhou2020perceppan}, and non-reference loss \citep{zhou2021pgman,luo2020pansharpening}.

\subsubsection{HS pansharpening}

Similar to pansharpening, HS pansharpening intends to combine spectral information in HS with spatial information in Pan to produce a HS image with high spatial resolution. 

\begin{itemize}
\item \textbf{Supervised methods} 
\end{itemize}

Supervised methods aim to learn the transformation from input to target data which do not exist in real world, thus simulation experiments are usually implemented. Specifically, a pair of low spatial resolution HS and low spectral resolution MS are generated by spatially and spectrally degrading the observed HS, respectively. By doing so, the two simulated images are regarded as the inputs of network and the original HS serves as reference.

Like those pioneer works in pansharpening, CNN and GAN are naturally applicable to HS pansharpening task. Representative single-branch CNN-based method \citep{zheng2020hyperspectral} proposes to utilize the channel-spatial-attention mechanism to adaptively extract informative features, where the DHP is used to upsample the input HS and can be enhanced by adding spatial-related constraints \citep{9664535}.
Residual structure is broadly utilized in two-branch HS pansharpening networks to learn the missing high frequency information. \cite{he2019hyperpnn} clearly exhibit the superiority of skip connection in terms of training efficiency. 
There are also enormous efforts aiming to tackle specific problem, such as spectral-fidelity \citep{he2020spectral,guan2021multistage}, pansharpening with arbitrary resolution enhancement \citep{he2021cnn} and arbitrary spectral bands \citep{qu2021dual}.
The hybrid networks such as DI-based methods \citep{dong2021laplacian} and VO-based methods \citep{xie2019hyperspectral} can adaptively learn the spatial details and deep priors that need explicit modeling by traditional methods.
Additionally, \cite{dong2021model} directly unfold the iterative optimization algorithm into a end-to-end network, where degradation models are considered to exploit the prior information. 
Following the idea presented in pansharpening, GAN is successfully applied to HS pansharpenging with various designs of the discriminators.
A typical example given by \cite{xie2020hpgan} utilizes a spatial discriminator to restrain the difference between input Pan and spectrally downsampled output of the generator, where the generator network is trained in the high frequency. Other commonly-used discriminators include the spectral discriminator \citep{dong2021fusion} and the spatial-spectral discriminator \citep{dong2021generative}.
Transformer also finds its application in HS pansharpening by \cite{bandara2022hypertransformer}, in which modality-specific feature extractor are designed to capture textural details for subsequent spectral details fusion.

\begin{itemize}
\item \textbf{Unsupervised methods} 
\end{itemize}

The unsupervised HS pansharpening is rarely studied compared with pansharpening. 
One possible reason is that the input Pan and MS share similar spectral coverage while there exisits a big discrepancy between Pan and HS in the spectral range, which leads to the difficulty in preserving spatial information.
A tentative work by \cite{nie2022unsupervised} utilizes a gradient and a high-frequency loss to model the spatial relationship, where an initialized image is first generated by the ratio estimation strategy.

\subsubsection{HS-MS data fusion}

Pansharpening related works can be regarded as special cases of HS-MS data fusion which aims to attain HS product with high spatial resolution by fusing paired HS-MS images. Therefore, many DL-based pansharpening methods can be transferred to tackle HS-MS fusion with necessary modifications. Following this, typical methods will be introduced in accordance with the same taxonomy in pansharpening.

\begin{itemize}
\item \textbf{Supervised methods} 
\end{itemize}


The supervised HS-MS fusion follows the same scheme of HS pansharpening by replacing the input Pan with MS.
Earlier DL-based HS-MS fusion methods are put forward with classic structures as 3-D CNN \citep{palsson2017multispectral}, residual network \citep{han2019deep}, dense connection network \citep{han2018ssf}, and three-component network \citep{zhang2020ssr}, etc. 
Compared with single-branch work that directly upsamples the HS to the same resolution as MS, multi-branch methods adopt an alternative strategy to relax this problem, i.e., by gradually upsampling the HS with deconvolution or pixel shuffle, where high resolution information in the corresponding scale are injected \citep{xu2020ham,han2019multi,zhou2019pyramid}. \cite {yang2018hyperspectral} use two branches to separately extract spectral-spatial features and then concatenate them for final reconstruction.
Recently, interpretable network combined with conventional models show great potential on this task, with examples either incorporate DI model into networks to adaptively learn detailed images \citep{sun2021band,lu2020coupled}, or design networks to automatically learn the observation models \citep{wang2021enhanced,wang2019deep} and deep priors \citep{dian2018deep,wang2021hyperspectral} in preparation for the subsequent fusion.
The deep unrolling methodology is also employed in HS-MS fusion, which effectively links the DL- and VO-based methods by unrolling the iterative optimization procedure into network training steps \citep{shen2021admm, xie2020mhf,xie2019multispectral,wei2020deep,yang2021variational}. 
Besides the prevalent CNN model, \cite{xiao2021physics} introduce a physical-based GAN method by embedding degradation models into the generator, where the output generated by degradation models are input into the discriminator for further spatial-spectral enhancement. Transformer is also introduced for HS-MS fusion \citep{hu2021fusformer}, where the structured embedding matrix is sent into transformer encoder to learn the residual map.

\begin{itemize}
\item \textbf{Unsupervised methods} 
\end{itemize}


Unsupervised HS-MS fusion methods only requires a pair of HS-MS images as the input of network and the fused HS can be obtained when the optimization of network is completed. These methods roughly comprise of two categories, i.e., encoding-decoding-based and generation-constraint-based methods. The former class assumes that the target image can be represented by multiplication of two matrices with each matrix standing for explicit physical meaning. AE is usually employed to model such procedure. The first work is proposed by \cite{qu2018unsupervised}, where the weights of decoder are shared by two AEs. Along this line, several successful methods sharing similar basic idea are proposed lately \citep{zheng2020coupled,yao2020cross,liu2022model}.
The other type of methods aim to directly generate the target image through a generator with an initialized image as input. In order to obtain a better reconstruction, extra information and constraints are needed to guide the parameter learning. The initialization can be the MS image at hand \citep{fu2019hyperspectral, han2019hyperspectral,li2022deep}, a random tensor \citep{uezato2020guided,liu2021umag}, and a specially learned code \citep{zhang2020deep,zhang2020unsupervised}.

\subsubsection{Spatiotemporal fusion}

Apart from the trade-off in spatial-spectral resolutions, there also exists a contradiction in spatial-temporal domain, i.e., images with high spatial resolution at the same area captured by current satellite platforms are usually obtained with a long time interval, and vice versa, which greatly hampers the practical applications such as change detection. Therefore, spatiotemporal fusion aims to produce temporally dense products with fine spatial resolution by fusing one or multiple pairs of coarse/fine images (e.g., MODIS-Landsat pairs) and a coarse spatial resolution image at the predicted time. This section introduces some typical methods in terms of their predicted land surface variables, e.g., reflectance, LST, NDVI, etc. 

A large majority of DL-based methods are designed for the reflectance images, where CNN prevail among all models. Inspired by the super-resolution problem, \cite{song2018spatiotemporal} propose the pioneering work, where a nonlinear mapping and a super-resolution network are learned to generate the predicted image. However, simply treating spatiotemporal fusion as a super-resolution problem inevitably impairs the performance due to the lack of the exploration in temporal information and hence many methods simultaneously exploiting the information underlying the spatial and temporal domains are proposed \citep{tan2018deriving, tan2019enhanced, li2020new}. Especially, \cite{liu2019stfnet} exploited temporal dependence and temporal consistent in the training process by incorporating the temporal information into the loss function, and hence obtain remarkable improvement. Compared with CNN, there are a few GAN-based methods that aim to generate outputs by optimizing a min-max problem. \cite{zhang2020remote} proposed a DL-based end-to-end trainable network in solving spatiotemporal fusion problem, where a two-stage framework are designed to gradually recover the predicted image. However, all the discussed methods require at least three images as inputs in the predicted stage, which may not be easily satisfied in practice. Thus, \cite{tan2021flexible} proposed a conditional GAN-based methods embedded with normalization techniques to eliminate the restriction on the number of input images. 

Compared with above models, DL-based methods originally designed for LST or NDVI are relatively scarce. Though some literature adopt reflectance-oriented methods to generate products of other land surface variables and obtain good performance, there still exist differences between these variables. Facing this problem, \cite{yin2020spatiotemporal} propose a LST-oriented methods by considering the temporal consistency, where two final outputs generated by multiscale CNN are fused together according to a novel weight function. As for NDVI products, \cite{jia2022multi} propose a multitask framework with a super-resolution net and a fusion net, where a time-constraint loss function is introduced to alleviate the time consistency assumption.

\subsection{Heterogeneous fusion}

Different from homogeneous fusion which aims to generate an outcome with high spectral, spatial, or temporal resolution based on pixel-level fusion, heterogeneous fusion mainly refers to the integration in LiDAR-optical, SAR-optical, RS-GBD, etc. Since the imaging mechanism of these data are totally different, feature-level and decision-level are widely adopted. 

\subsubsection{LiDAR-optical fusion}
LiDAR-optical fusion can be applied to many tasks, e.g., registration, pansharpening, target extraction, estimation of forest biomass \citep{zhang2017advances}. Since it is hard to give a thorough and detailed introduction concerning all aspects, we focus on one particular domain, i.e., HS-LiDAR data fusion in the application of LULC classification, and give some examples employed in other tasks.

HS data has been widely used in classification task by virtue of its rich spectral information, but the performance inevitably meets the bottleneck in the situation where spectral information is not sufficient to discriminate the targets. Luckily, the LiDAR system is capable of acquiring 3-D spatial geometry, which compensates for the shortage in HS, and hence the joint utilization of HS and LiDAR data in identifying materials becomes a hot spot in recent years. \cite{ghamisi2016hyperspectral} pioneer the first DL-based HS-LiDAR fusion network, where features of input data are extracted by EPs and then integrated by two fusion strategies for the consequent DL-based classifier. Though great improvement is achieved compared with traditional methods, the way in feature extraction and feature fusion is simple and rough, which limits further improvements to some extent. Inspired by this milestone, many advanced methods have been proposed, aiming at improving the two critical steps. For the feature extraction, a typical example is given by \cite{chen2017deep} who utilize a two-branch network to separately extract spectral-spatial-elevation features and then a fully connected layer is used to integrate these heterogeneous features for final classification. Other particularly designed features extraction networks include a three-branch network \citep{li2018hyperspectral}, a dual-tunnel network \citep{xu2017multisource, zhao2020joint}, and a encoder–decoder translation network \citep{zhang2018feature}. For the feature fusion, \cite{feng2019multisource} incorporate Squeeze-and-Excitation networks into the fusion step to adaptively realize feature calibration. Other novel fusion strategies are also proposed, such as cross-attention module \citep{mohla2020fusatnet}, a reconstruction-based network \citep{hong2020deep}, a feature-decision combined fusion network \citep{hang2020classification}, and a graph fusion network \citep{du2021multisource}.

Researchers in LiDAR-optical fusion also pay attention to target extraction, such as buildings, roads, impervious surfaces, etc. \cite{huang2019automatic} propose a encoder-decoder network embedded with a gated feature labeling unit to identify the buidings and non-buildings areas. Algorithms in extracting roads and impervious surfaces are also proposed by \cite{parajuli2018fusion} and \cite{sun2019extracting}, respectively. Very recent, \cite{han2022multimodal} propose the first DL-based multimodal unmixing network, where the height information from LiDAR extracted by the squeeze-and-excitation attention module is used to guide the unmixing process in HS.

\subsubsection{SAR-optical data fusion}

Different from optical images, SAR system is designed to collect backscatter signals of ground objects that can not only reflect the information of RADAR system parameters but also embody the physical and geometric characteristics of the observed scenes.
Although SAR data can provide complementary knowledge for optical images, it is highly prone to speckle noise that may heavily restrict its practical potential. The joint use of SAR and optical data becomes a feasible solution to realize better understanding and analysis of targets of interest. 

According to which level the fusion is carried out, we can divide SAR-optical data fusion into three categories, namely, pixel-level, feature-level, and decision-level. Though there exists a large gap between SAR and optical data in the imaging mechanism, it is feasible to synthetically generate an optical product with abundant textural and structural information with the aid of SAR image through a pixel-level fusion. In that case, registration becomes extremely crucial and many DL-based registration methods between SAR and optical data are proposed, such as the siamese CNN \citep{zhang2019registration}, and the self-learning and transferable network \citep{wang2018deep}.
After obtaining a pair of co-registered  SAR-optical data, many traditional methods originally designed for pansharpening are extended for the SAR-optical pixel-level fusion. \cite{kong2021fusion} propose a GAN-based network containing a U-shaped generator and a convolutional discriminator, where extensive losses are taken into consideration to fully eliminate the speckle noise and preserve abundant structure information. 
In addition, optical images are easily subject to atmospheric conditions, where the cloud cover critically impair the spectral and spatial information. Luckily, SAR is almost insensitive to these factors thanks to its independence from weather conditions. Thus, many pixel-level-based methods are designed to generate a cloud-free optical image from the corresponding cloud-corrupted optical image with the help of an auxiliary SAR data at the same area. \cite{meraner2020cloud} adopt a simple residual structure to directly learn the mapping from the input data pairs to the cloud-free target and demonstrate its superiority even in the situation where scenes are covered by thick clouds. Besides, GAN-based methods are also proposed to remove clouds \citep{gao2020cloud,grohnfeldt2018conditional}.

In addition to pixel-level fusion, high level fusion for applications like LULC classification also catches considerable interest using SAR-optical data. \cite{hu2017fusionet} propose the first DL-based HS-SAR data fusion network, in which a simple yet effective two-branch architecture is used to separately extract heterogeneous features for final convolutional fusion. Nevertheless, the efficiency of such straightforward strategy of feature extraction remains limited without considering information redundancy. Hence, a novel BN technique constrained by the sparse constraint is devised to reduce the unnecessary features and make the network generalize better \cite{li2022asymmetric}. In addition to the tasks mentioned above, SAR-optical fusion has also been applied to change detection \citep{li2021deep}, biomass estimation \citep{shao2017stacked}, etc.

\subsubsection{RS-GBD fusion}

GBD contain a wide range of sources from social media, geographic information systems, mobile phones, etc, which greatly contribute to the understanding in our living environment. More specifically, RS exhibit a strong ability in capturing physical attributes of a large-scale earth surface from a global view. On the other hand, the information provided by GBD is highly associated with human behaviors, which gives abundant socioeconomic descriptions as a supplement to RS. Notably there exists a big gap between GBD and RS in the data structures, therefore current popular dual-branch network that is widely used to extract modality-specific features can not be directly employed to the fusion of GBD and RS data. This section sort out some successful examples in RS-GBD fusion according to the category of GBD used in the fusion process, such as street view imagery, POI, vehicle trajectory data, etc.


POI refers to the objects that can be abstracted into a point, such as theaters, bus stops, and houses. Different from RS data, each POI generally contains name, coordinate and some other geographic information, which can be easily gleaned by electronic maps, such as OpenStreetMap. Since the attributes of each POI have close correlation with functional facilities, the integration between POI and RS poses a new opportunity to the task in urban functional zone classification. Very recently, \cite{lu2022unified} propose a unified DL-based method to jointly exploit characteristic features underlying POI and RS. Concretely, POI are firstly converted into a distance heatmap to meet the input requirement of CNN, and then two modules are used for feature extraction and spatial relation exploration respectively. Other related algorithms with different structures are proposed, such as a deep multi-scale network \citep{xu2020new,bao2020dfcnn} and a bi-branch network \cite{fan2021urban}.

In addition to POI, street view imagery is one important data source that can be gathered from social media (e.g., Twitter, Instagram, and Weibo) and street view cars (e.g., Google, Baidu, and Gaode). Different from RS data, it gives fine-grained pictures along the street networks from human's view, and hence provides a diverse and complementary descriptions about our surroundings \citep{7926323}. A typical example is given by \cite{srivastava2019understanding} who utilize the RS and Google street view data to realize urban land use classification. More concretely, a two-branch structured network is used to separately extract features from both modalities which are then stacked into new feature for later classification. It is worth mentioning that authors propose a novel solution to deal with the tricky situation where one modality data is missing during the testing phase.

Besides, other kinds of GBD are also widely used in the fusion task. Taxi trajectory data and user visit data are utilized to identify the urban functional areas \citep{qian2020identification,cao2020deep}. \cite{mantsis2020multimodal} use the snow-related twitters along with Sentinel-1 images to realize snow depth estimation. \cite{lu2022unified} employ a two-branch network to extract information from RS and Tencent user density to estimate the proportion of mixed land use. 

\section{List of resources} \label{List of resources}

With a massive number of multimodal RS data available, DL-based technologies have witnessed considerable breakthroughs in data fusion. Numerous DL models and related algorithms using various multimodal data are springing up, which provides endless inspirations for people who take up the research on DL-based multimodal RS data fusion. For the sake of developments and communications on this domain, we collect and summarize some relevant resources, including tutorials for the beginners, available multimodal RS data used in the literature, and open-source codes provided by the authors.

\subsection{Tutorials}

We further give some materials and references for beginners who are willing to work on DL-related RS tasks, as listed in table \ref{tab:tutorials}. The references in the category of RS can give readers a quick and comprehensive view in the features, principles and applications of different modalities from RS. Materials in DL introduce some widely-used models that constitute the pillars of almost all DL-based algorithms. Following that, we recommend five classic references in RS \& DL, which aims to present some successful applications in RS achieved by DL. From the aforementioned tutorials along with their citations, readers can have a basic knowledge of relevant backgrounds in preparation for the further research.

\begin{table}[!t]
\centering
\caption{Some tutorials for beginners}
\label{tab:tutorials}
\begin{adjustbox}{width=1\textwidth}
\begin{tabular}{ccll}
\hline
\multicolumn{1}{l}{}       & \multicolumn{1}{l}{Aspects} & References                                                                                              & Descriptions                                                                                 \\ \hline
\multirow{11}{*}{Tutorials} & \multirow{3}{*}{RS}         & \citep{bioucas2013hyperspectral}                                         & Introducing basic concepts and features of HS and its relevant topics                        \\ \cline{3-4} 
                           &                             & \citep{rasti2020feature}                     & Providing a review of feature extraction approaches in HS           \\ \cline{3-4} 
                           &                             & \citep{moreira2013tutorial}                                                                 & Giving principles and theories of SAR and its techniques and applications                    \\ \cline{2-4} 
                           & \multirow{3}{*}{DL}         & \citep{schmidhuber2015deep}                                                      & Reviewing deep supervised learning, unsupervised learning, and reinforcement learning        \\ \cline{3-4} 
                           &                             & \citep{liu2017survey}                                  & Introducing typical DL architectures and their applications                                  \\ \cline{3-4} 
                           &                             & \citep{zhang2018survey}                                                                & Reviewing DL models and their applications in analysising big data                           \\ \cline{2-4} 
                           & \multirow{5}{*}{RS \& AI}   & \citep{zhang2019remotely} & Introducing three main development stages for RS and focusing on DL for RS big data          \\ \cline{3-4} 
                           &                             & \citep{zhang2016deep}              & Introducing typical DL models and their applications in RS tasks                             \\ \cline{3-4} 
                           &                             & \citep{zhu2017deep}              & Reviewing DL models and related algorithms in RS domains followed by a list of resources     \\ \cline{3-4} 
                           &                             & \citep{hong2021interpretable}              & Giving a survey of nonconvex modeling toward interpretable AI models in HS     \\ \cline{3-4}

                           &                             & \citep{ma2019deep}                              & Conducting a literature survey by meta-analysis method and introducing relevant applications \\ \hline
\end{tabular}
\end{adjustbox}
\end{table}


\subsection{Available multimodal RS data}

To comprehensively evaluate the existing algorithms and select suitable models for practical applications, available multimodal RS datasets are indispensable taches of the whole fusion procedure. Thanks to the Data Fusion Technical Committee (DFTC) of the IEEE Geoscience and Remote Sensing Society, a Data Fusion Contest is held annually since 2006, which provides researchers valuable multimodal RS datasets and promotes the development in data fusion domain. Nowadays, these available datasets have been widely used in the literature for the methodology evaluation. More information can be found in \citep{dalla2015challenges} and \citep{kahraman2021comprehensive} which provide a detailed summary of these datasets and their applications. Hence, in this section we collect available datasets except the aforementioned datasets provided by DFTC, contributing to the RS community.

\begin{table}[!t]
\centering
\caption{Non-exhaustive list of multimodal RS datasets}
\label{tab:datasets}
\begin{adjustbox}{width=1.0\textwidth}
\begin{tabular}{cllll}
\hline
\multicolumn{1}{l}{}         & Source   & Reference & Descriptions   & Link                     \\ \hline
\multirow{2}{*}{Pansharpening}        & \begin{tabular}[c]{@{}l@{}}Ikonos, QuickBird, Gaofen-1,\\ and WorldView-2/3/4\end{tabular} & \citep{meng2020large}         
                                       & \begin{tabular}[c]{@{}l@{}}2,270 pairs of HR Pan/LR MS images from different \\kinds of remote sensing satellites\end{tabular}          & http://www.escience.cn/people/fshao/database.html                  \\ \cline{2-5} 
                     
                                      & \begin{tabular}[c]{@{}l@{}}GeoEye-1,WorldView-2/3/3,\\and SPOT-7,Pléiades-1B \end{tabular} & \citep{vivone2021benchmarking}         
                                       & \begin{tabular}[c]{@{}l@{}}14 pairs of Pan-MS images collected \\over heterogeneous landscapes by different satellites \end{tabular}
                                      & https://resources.maxar.com/product-samples/pansharpening-benchmark-dataset                        \\ \hline
                    
             HS pensharpening                      & PRISMA              & None         
                                       & 4 pairs of HR Pan/LR HS images provided by WHISPER     
                                       & https://openremotesensing.net/hyperspectral-pansharpening-challenge/                              \\ \hline
                    
\multirow{2}{*}{Spatiotemporal}       & Landsat8, MODIS          & \citep{li2020spatio}        
                                      & \begin{tabular}[c]{@{}l@{}}27,27, and 29 pairs of Landsat-MODIS images from 3\\ different datasets       \end{tabular}                 
                                      & https://drive.google.com/open?id=1yzw-4TaY6GcLPIRNFBpchETrFKno30he \\ \cline{2-5} 
                   
                                      & Landsat5/7, MODIS        & \citep{emelyanova2013assessing}         
                                      & 14 and 17 pairs of Landsat-MODIS images from 2 dataset                                     
                                      & https://data.csiro.au/collection/csiro:5846 and https://data.csiro.au/collection/csiro:5847                                                             \\ \hline
                                      
\multirow{4}{*}{LULC classification}  & Sentinel-2, ITRES CASI-1500  & \citep{hong2021multimodal}         
                                      & \begin{tabular}[c]{@{}l@{}}HS-MS scene with 349 × 1905 pixels covering the University \\of Houston\end{tabular}                           & https://github.com/danfenghong/ISPRS\_S2FL                                                 \\ \cline{2-5} 
                                     
                                      & EnMAP, Sentinel-1                            & \citep{hong2021multimodal}        
                                      & \begin{tabular}[c]{@{}l@{}}HS-SAR scene with 1723 × 476 pixels covering the Berlin\\ urban and its neighboring area  \end{tabular}         & https://github.com/danfenghong/ISPRS\_S2FL                                                        \\ \cline{2-5} 
                                      
                                      & \begin{tabular}[c]{@{}l@{}}HySpex, Sentinel-1, and \\DLR-3 K system \end{tabular}          & \citep{hong2021multimodal}         
                                      & HS-SAR-DSM with 332 × 485 pixels over Augsburg                                               & https://github.com/danfenghong/ISPRS\_S2FL                                                      \\ \cline{2-5} 
                                      
                                      & ITRES CASI-1500, ALTM                        & \citep{gader2013muufl}         
                                      & \begin{tabular}[c]{@{}l@{}}HS-LiDAR with 325 × 220 pixels over the University of \\ Southern Mississippi Gulf Park Campus  \end{tabular} & https://github.com/GatorSense/MUUFLGulfport/tree/master/MUUFLGulfportSceneLabels               \\ \hline
                                      
\multirow{3}{*}{Objection extraction} & \begin{tabular}[c]{@{}l@{}}USGS, OSM, state, and \\ federal agencies \end{tabular}       & \citep{huang2019automatic}      
                                      & \begin{tabular}[c]{@{}l@{}}Orthophotos, LiDAR point clouds, and ground-truth building \\ masks       \end{tabular}                      
                                      & https://dx.doi.org/10.6084/m9.figshare.3504413                                                    \\ \cline{2-5} 
                                      
                                      & Commission II/4 of the ISPRS                 & \citep{hosseinpour2022cmgfnet}        
                                      & Orthophotos with corresponding DSM and labels                                               
                                      & https://www.isprs.org/education/benchmarks/UrbanSemLab/semantic-labeling.aspx                     \\ \cline{2-5} 
                                      
                                      & TLCGIS                                   & \citep{parajuli2018fusion}        
                                      & RGB images, LiDAR-derived depth images, and road masks                                         & https://bitbucket.org/biswas/fusion\_lidar\_images/src/master/                                      \\ \hline
\end{tabular}
\end{adjustbox}
\end{table}


\subsection{Open-source codes in DL-based multimodal RS data fusion}

For the researchers who already have some background knowledge in this domain and are ready to design their own algorithms, the open-source codes can provide them tremendous help. In that case, we search and summarize available codes from GitHub and authors' homepages in table \ref{tab:code} for the sake of comparison between different approaches.


\begin{table}[!htbp]
\centering
\caption{Open-source codes in DL-based multimodal RS data fusion}
\label{tab:code}
\begin{adjustbox}{width=0.83\textwidth}
\begin{tabular}{ccllll}
\hline
\multicolumn{1}{l}{}              & \multicolumn{1}{l}{Categories}        & Name                    & References & Languages/Frameworks & Links                         \\ \hline
\multirow{23}{*}{Pansharpening}   & \multirow{18}{*}{Supervised}          & A-PNN                   & \citep{scarpa2018target}          & Theano               & https://github.com/sergiovitale/pansharpening-cnn-python-version                                    \\ \cline{3-6} 
                                  &                                       & PanNet                  & \citep{yang2017pannet}            & Chainer              & https://github.com/oyam/PanNet-Landsat                                                              \\ \cline{3-6} 
                                  &                                       & PNN                     & \citep{masi2016pansharpening}           & MATLAB               & http://www.grip.unina.it/research/85-image-enhancement/93-pnn.html                                  \\ \cline{3-6} 
                                  &                                       & GTP-PNet                & \citep{zhang2021gtp}           & TensorFlow           & https://github.com/HaoZhang1018/GTP-PNet                                                            \\ \cline{3-6} 
                                  &                                       & SDPNet                  &   \citep{xu2020sdpnet}       & TensorFlow           & https://github.com/hanna-xu/SDPNet-for-pansharpening                                                \\ \cline{3-6} 
                                  &                                       & TFNet                   &   \citep{liu2020remote}         & PyTorch              & https://github.com/liouxy/tfnet\_pytorch                                                            \\ \cline{3-6} 
                                  &                                       & DiCNN                   &   \citep{he2019pansharpening}         & TensorFlow           & https://github.com/whyLemon/Pansharpening-via-Detail-Injection-Based-Convolutional-Neural-Networks- \\ \cline{3-6} 
                                  &                                       & Fusion-Net              &  \cite{deng2020detail}          & TensorFlow           & https://github.com/liangjiandeng/FusionNet                                                          \\ \cline{3-6} 
                                  &                                       & TDNet                   &  \citep{9732243}          & PyTorch              & https://github.com/liangjiandeng/TDNet                                                              \\ \cline{3-6} 
                                  &                                       & VO+Net                  & \citep{wu2021vo+}           & MATLAB               & https://github.com/liangjiandeng/VOFF                                                               \\ \cline{3-6} 
                                  &                                       & VP-Net                  &   \citep{tian2021vp}         & TensorFlow           & https://github.com/likun97/VP-Net                                                                   \\ \cline{3-6} 
                                  &                                       & DL-VM                   &  \citep{shen2019spatial}          & MATLAB               & https://github.com/WHU-SGG-RS-Pro-Group/DL\_VM                                                      \\ \cline{3-6} 
                                  &                                       & MDSSC-GAN               &   \citep{gastineau2021generative}         & TensorFlow           & https://github.com/agastineau/MDSSC-GAN\_SAM                                                        \\ \cline{3-6} 
                                  &                                       & PanColorGAN             &  \citep{ozcelik2020rethinking}          & PyTorch              & https://github.com/ozcelikfu/PanColorGAN                                                            \\ \cline{3-6} 
                                  &                                       & PSGAN                   & \citep{liu2020psgan}           & PyTorch              & https://github.com/zhysora/PSGan-Family                                                             \\ \cline{3-6} 
                                  &                                       & RED-cGAN                &  \citep{shao2019residual}          & TensorFlow           & https://github.com/Deep-Imaging-Group/RED-cGAN                                                      \\ \cline{3-6} 
                                  &                                       & ArbRPN                  &    \citep{chen2021arbrpn}        & PyTorch              & https://github.com/Lihui-Chen/ArbRPN                                                                \\ \cline{3-6} 
                                  &                                       & PanFormer               &  \citep{zhou2022panformer}          & PyTorch              & https://github.com/zhysora/PanFormer                                                                \\ \cline{2-6} 
                                  & \multirow{5}{*}{Unsupervised}         & UCGAN                   &  \citep{zhou2022unsupervised}          & PyTorch              & https://github.com/zhysora/UCGAN                                                                    \\ \cline{3-6} 
                                  &                                       & Pan-GAN                 &  \citep{ma2020pan}          & TensorFlow           & https://github.com/yuwei998/PanGAN                                                                  \\ \cline{3-6} 
                                  &                                       & PercepPan               &  \citep{zhou2020perceppan}          & PyTorch              & https://github.com/wasaCheney/PercepPan                                                             \\ \cline{3-6} 
                                  &                                       & PGMAN                   &  \citep{zhou2021pgman}          & PyTorch              & https://github.com/zhysora/PGMAN                                                                    \\ \cline{3-6} 
                                  &                                       & ZeRGAN                  &   \citep{diao2022zergan}         & PyTorch              & https://github.com/RSMagneto/ZeRGAN                                                                 \\ \hline
\multirow{7}{*}{HS pansharpening} & \multirow{7}{*}{Supervised}           & DIP-HyperKite           &  \citep{9664535}          & PyTorch              & https://github.com/wgcban/DIP-HyperKite                                                             \\ \cline{3-6} 
                                  &                                       & DBDENet                 &  \citep{qu2021dual}          & PyTorch              & https://github.com/Jiahuiqu/DBDENet/tree/111528a82e579faaf02d4ffd3ea0df0e51de2efb                   \\ \cline{3-6} 
                                  &                                       & MDA-Net                 &  \citep{guan2021multistage}          & PyTorch              & https://github.com/pyguan88/MDA-Net                                                                 \\ \cline{3-6} 
                                  &                                       & MSSL                    &  \citep{qu2021mssl}          & PyTorch              & https://github.com/Jiahuiqu/MSSL                                                                    \\ \cline{3-6} 
                                  &                                       & MoG-DCN                 &  \cite{dong2021model}          & PyTorch              & https://github.com/chengerr/Model-Guided-Deep-Hyperspectral-Image-Super-resolution                  \\ \cline{3-6} 
                                  &                                       & Pgnet                   &   \citep{li2022unmixing}         & PyTorch              & https://github.com/rs-lsl/Pgnet                                                                     \\ \cline{3-6} 
                                  &                                       & HyperTransformer        &   \citep{bandara2022hypertransformer}         & PyTorch              & https://github.com/wgcban/HyperTransformer                                                          \\ \hline
\multirow{27}{*}{HS-MS}           & \multirow{14}{*}{Supervised}          & SSR-NET                 &  \citep{zhang2020ssr}          & PyTorch              & https://github.com/hw2hwei/SSRNET                                                                   \\ \cline{3-6} 
                                  &                                       & HSRnet                  & \citep{hu2021hyperspectral}           & TensorFlow           & https://github.com/liangjiandeng/HSRnet                                                             \\ \cline{3-6} 
                                  &                                       & PZRes-Net               & \citep{zhu2020hyperspectral}           & PyTorch              & https://github.com/zbzhzhy/PZRes-Net                                                                \\ \cline{3-6} 
                                  &                                       & Two-CNN-Fu              &  \citep{yang2018hyperspectral}          & Caffe                & https://github.com/polwork/Hyperspectral-and-Multispectral-fusion-via-Two-branch-CNN                \\ \cline{3-6} 
                                  &                                       & ADMM-HFNet              &  \citep{shen2021admm}          & TensorFlow           & https://github.com/liuofficial/ADMM-HFNet                                                           \\ \cline{3-6} 
                                  &                                       & MHF-Net                 &   \citep{xie2020mhf}         & TensorFlow           & https://github.com/XieQi2015/MHF-net                                                                \\ \cline{3-6} 
                                  &                                       & CNN-Fus                 &    \citep{dian2020regularizing}        & MATLAB               & https://github.com/renweidian/CNN-FUS                                                               \\ \cline{3-6} 
                                  &                                       & DHIF-Net                &  \citep{huang2022deep}          & PyTorch              & https://github.com/TaoHuang95/DHIF-Net                                                              \\ \cline{3-6} 
                                  &                                       & DHSIS                   &  \citep{dian2018deep}          & MATLAB+Keras         & https://github.com/renweidian/DHSIS                                                                 \\ \cline{3-6} 
                                  &                                       & EDBIN                   &   \citep{wang2021enhanced}         & TensorFlow           & https://github.com/wwhappylife/Deep-Blind-Hyperspectral-Image-Fusion                                \\ \cline{3-6} 
                                  &                                       & SpfNet                  & \citep{liu2022patch}           & TensorFlow           & https://github.com/liuofficial/SpfNet                                                               \\ \cline{3-6} 
                                  &                                       & TONWMD                  &  \citep{shen2020twice}          & TensorFlow           & https://github.com/liuofficial/TONWMD                                                               \\ \cline{3-6} 
                                  &                                       & TSFN                    &   \citep{wang2021hyperspectral}         & MATLAB+PyTorch       & https://github.com/xiuheng-wang/Sylvester\_TSFN\_MDC\_HSI\_superresolution                          \\ \cline{3-6} 
                                  &                                       & Fusformer               &   \citep{hu2021fusformer}         & PyTorch              & https://github.com/J-FHu/Fusformer                                                                  \\ \cline{2-6} 
                                  & \multirow{13}{*}{Unsupervised}        & CUCaNet                 &   \citep{yao2020cross}         & PyTorch              & https://github.com/danfenghong/ECCV2020\_CUCaNet                                                    \\ \cline{3-6} 
                                  &                                       & HyCoNet                 & \citep{zheng2020coupled}           & PyTorch              & https://github.com/saber-zero/HyperFusion                                                           \\ \cline{3-6} 
                                  &                                       & MIAE                    &\citep{liu2022model}            & PyTorch              & https://github.com/liuofficial/MIAE/tree/c880d1df15f022f78fc8305e436aa0bfae378135                   \\ \cline{3-6} 
                                  &                                       & NonRegSRNet             &  \citep{zheng2021nonregsrnet}          & PyTorch              & https://github.com/saber-zero/NonRegSRNet                                                           \\ \cline{3-6} 
                                  &                                       & u$^2$-MDN                 &  \citep{qu2021unsupervised}          & TensorFlow           & https://github.com/yingutk/u2MDN                                                                    \\ \cline{3-6} 
                                  &                                       & uSDN                    &   \citep{qu2018unsupervised}         & TensorFlow           & https://github.com/aicip/uSDN                                                                       \\ \cline{3-6} 
                                  &                                       & HSI-CSR                 &  \citep{fu2019hyperspectral}          & Caffe                & https://github.com/ColinTaoZhang/HSI-SR                                                             \\ \cline{3-6} 
                                  &                                       & DBSR                    &  \citep{zhang2020deep}          & PyTorch              & https://github.com/JiangtaoNie/DBSR                                                                 \\ \cline{3-6} 
                                  &                                       & GDD                     &   \citep{uezato2020guided}         & PyTorch              & https://github.com/tuezato/guided-deep-decoder                                                      \\ \cline{3-6} 
                                  &                                       & UAL                     &   \citep{zhang2020unsupervised}         & PyTorch              & https://github.com/JiangtaoNie/UAL-CVPR2020                                                         \\ \cline{3-6} 
                                  &                                       & UDALN                   &  \citep{li2022deep}          & PyTorch              & https://github.com/JiaxinLiCAS/UDALN\_GRSL                                                          \\ \cline{3-6} 
                                  &                                       & RAFnet                  &   \citep{lu2020rafnet}         & TensorFlow           & https://github.com/RuiyingLu/RAFnet                                                                 \\ \cline{3-6} 
                                  &                                       & Rec\_HSISR\_PixAwaRefin &  \citep{wei2020unsupervised}          & PyTorch              & https://github.com/JiangtaoNie/Rec\_HSISR\_PixAwaRefin                                              \\ \hline
\multirow{3}{*}{Spatiotemporal}   & \multirow{2}{*}{CNN}                  & DCSTFN                  &  \citep{tan2018deriving}           & TensorFlow           & https://github.com/theonegis/rs-data-fusion                                                         \\ \cline{3-6} 
                                  &                                       & EDCSTFN                 &  \citep{tan2019enhanced}         & PyTorch              & https://github.com/theonegis/edcstfn                                                                \\ \cline{2-6} 
                                  & GAN                                   & GANSTFM                 &   \citep{tan2021flexible}         & PyTorch              & https://github.com/theonegis/ganstfm                                                                \\ \hline
\multirow{14}{*}{LiDAR-optical}   & \multirow{11}{*}{LULC classification} 
                                  & AM$^3$Net                  &  \citep{wang2022am3net}          & PyTorch              &                                                                                  https://github.com/Cimy-wang/AM3Net\_Multimodal\_Data\_Fusion          \\ \cline{3-6} 
                                  &                                       & CCR-Net                 &  \citep{wu2021convolutional}          & TensorFlow           & https://github.com/danfenghong/IEEE\_TGRS\_CCR-Net                                                  \\ \cline{3-6} 
                                  &                                       & EndNet                  & \citep{hong2020deep}           & TensorFlow           & https://github.com/danfenghong/IEEE\_GRSL\_EndNet                                                   \\ \cline{3-6} 
                                  &                                       & FusAtNet                &  \citep{mohla2020fusatnet}          & Keras                & https://github.com/ShivamP1993/FusAtNet                                                             \\ \cline{3-6} 
                                  &                                       & HRWN                    &  \citep{zhao2020joint}          & Keras                & https://github.com/xudongzhao461/HRWN                                                               \\ \cline{3-6} 
                                  &                                       & IP-CNN                  &   \citep{zhang2021information}         & Keras                & https://github.com/HelloPiPi/IP-CNN-code                                                            \\ \cline{3-6} 
                                  &                                       & MAHiDFNet               & \citep{wang2022multi}           & Keras                & https://github.com/SYFYN0317/-MAHiDFNet                                                             \\ \cline{3-6} 
                                  &                                       & MDL-RS                  &   \citep{hong2021more}         & TensorFlow           & https://github.com/danfenghong/IEEE\_TGRS\_MDL-RS                                                   \\ \cline{3-6} 
                                  &                                       & RNPRF-RNDFF-RNPMF       & \citep{ge2021deep}           & Keras                & https://github.com/gechiru/RNPRF-RNDFF-RNPMF                                                        \\ \cline{3-6} 
                                  &                                       & S²ENet                  &\citep{fang2021s2enet}            & PyTorch              & https://github.com/likyoo/Multimodal-Remote-Sensing-Toolkit                                         \\ \cline{3-6} 
                                  &                                       & two-branch CNN          &   \citep{xu2017multisource}         & Keras                & https://github.com/Hsuxu/Two-branch-CNN-Multisource-RS-classification                               \\ \cline{2-6} 
                                  
                                  & \multirow{2}{*}{Target extraction}    & CMGFNet                  &  \citep{hosseinpour2022cmgfnet}          & PyTorch              &
                                  https://github.com/hamidreza2015/CMGFNet-Building\_Extraction  \\ \cline{3-6} 
                                  &                                       & GRRNet                   &  \citep{huang2019automatic}               & Caffe              &
                                   https://github.com/CHUANQIFENG/GRRNet   \\ \cline{2-6} 
                                  
                                  & Unmixing                              & MUNet                   &   \citep{han2022multimodal}         & PyTorch              & https://github.com/hanzhu97702/IEEE\_TGRS\_MUNet                                     \\ \hline
                                  
\multirow{2}{*}{RS-GBD}           & POI data                              &  UnifiedDL-UFZ                & \citep{lu2022unified}           & PyTorch                & https://github.com/GeoX-Lab/UnifiedDL-UFZ-extraction                                                    \\ \cline{2-6} 
                                  & User density data                     &  CF-CNN           &  \citep{he2020accurate}          & Keras              &     https://github.com/SysuHe/MultiSourceData\_CFCNN                                            \\ \hline

\end{tabular}
\end{adjustbox}
\end{table}


\section{Problems and prospects}\label{Problems and prospects}

A great deal of progress has been made recently in the DL-based multimodal RS data fusion. However, there still exists some problems remaining to be solved. This section aims to point out current challenges faced by the fast-growing domain and present prospects for the future directions.

\subsection{From well-registered to non-registered} 
Image registration is a fundamental prerequisite for many RS tasks, such as data fusion and change detection. Since the accuracy of registration between two modalities has a non-negligible influence on the image fusion, aligning to-be-fused data with high precision become an extremely important step before the fusion process, especially for the pixel-level fusion. It is not a difficult task for the fusion between Pan and MS because there are many platforms equipped with the two modality sensors, and hence paired Pan-MS images are easily obtained under the same atmospheric environment and at the same acquisition time, which greatly reduce the registration difficulty. It is rather harder to obtain paired HS-Pan or HS-MS images under the same situation, so data registration becomes a crucial task compare with Pan-MS. However, much attention has been paid to designing advanced algorithms in the DL-based multimodal RS data fusion domain by assuming that the input data are perfectly co-registered, and thus ignoring the importance of such preprocessing. Only a few DL-based fusion work focus on the multitask by jointly realizing image registration and fusion. Very recently, \cite{zheng2021nonregsrnet} make an attempt to realize the registration and fusion tasks in an end-to-end unsupervised fusion network, where the inputs are a pair of unregistered HS-MS data. In the future, it is advisable to pay more attention to the registration step and incorporate this preprocessing into the fusion process.

\subsection{From image-oriented to application-oriented quality assessment}

Quality assessment for the output product is an indispensable part of the whole fusion process. The evaluation for high level fusion, i.e., feature-level and decision-level, generally depends on the performance of subsequent applications, such as classification, target detection, and change detection. However, the assessment for pixel-level fusion is usually implemented by calculating related indexes from spatial and spectral domains, and it can be divided into two categories, i.e., quality with reference and quality with no reference. For the first class, some widely-used indexes, such as SSIM, SAM, and ERGAS, are calculated between the fused product and the reference image. However, the existing indexes are not sufficient enough to exhibit and compare various methods in a comprehensive and fair way, which inevitably hinder users from selecting appropriate methods for real-world applications. Very recently, \cite{zhu2022novel} propose a novel framework for the quality assessment of spatiotemporal products, which not only takes the spatial and spectral errors into consideration but also the characteristics of input data and land surface. On the other hand, it is very likely that the reference images are not readily available in practice, so designing an index without the requirement of the reference is urgently needed. \cite{liu2015human} proposed a non-reference index for the pansharpening by using Gaussian scale space which is more consistent with human visual system. Besides, some researchers adopt application-oriented evaluating indicators to judge the performance of pansharpening methods, for example, \cite{qu2017does} evaluate the pansharpening approaches by comparing anomaly detection performance in their pansharpened outputs. In general, it is more desirable to employ application-related indexes to evaluate different algorithms since the purpose of fusion is to combine complementary information for a better decision in a specific application. Therefore, it is a good way for DL-based fusion methods to incorporate the application-related indexes into theirs loss functions to guide the network to learn representative outputs which are more suitable for the subsequent applications.

\subsection{From two-modality to multi-modality}
With the quick development of diverse sensors on airborne and spaceborne platforms, the availability of modalities becomes more diverse. Currently, most of DL-based fusion algorithms are designed for only two-modality, limiting the application ability of multi-modality. As a result, \emph{how to effectively utilize more modality data and fully exhibit their potentials as well as further make the performance bottleneck is a remaining challenge in the multimodal data fusion task.} More importantly, with more and more modality data easily accessible, future researches could consider developing a unified DL-based framework which could deal with arbitrary number of modalities as the inputs.

\subsection{From multimodal to crossmodal learning} 
Though multimodal data with diverse features contribute to our understanding in the world, it is more likely that some modality data is absent in practical scenarios. For example, SAR and MS data are available on a global scale. In contrast, HSI data are more hard to collect on account of the limitation of sensors, which may lead to the shortage in some areas. Hence, \emph{how to transfer the information hidden in the area with multimodal data into the scenario where one modality is missing} is a typical issue that crossmodal learning aims to deal with. A representative DL-based method tackling this practical problem is proposed by \citep{hong2020x}, where a limited number of HS-MS or HS-SAR pairs are used in the training phase to realize a large-scale classification task in an area only covered by one modality data, i.e., MS or SAR. In the future, it is believed that this critical domain will catch more attention in the RS fusion community under the influence of RS big data and DL.

\subsection{From black-box to interpretable DL}
Though DL has witnessed numerous breakthroughs in recent years, it is often accused of an inexplicable black-box learning procedure. Unlike the traditional methods which has clear physical and mathematical meanings, DL-based methods extract high-level features which are hard to explain. As discussed in section \ref{Pansharpening}, 
many model-driven DL-based methods are successively proposed to design a totally interpretable networks with each module presenting a specific operation. The combination between model-driven and data-driven methods poses a new view to understand the workflow in the black-box network and also points out a solution to make the black-box transparent. However, this solution is limited to spatiospectral fusion domain and hard to apply to feature-level and decision-level fusion. Therefore, high-level fusion still stays at the stage where researchers pay much attention to the feature extraction and feature fusion instead of fully understanding what the network really learns. However, understanding the features learned by each hidden layer contributes to designing more effective network structures in mining discriminative features and hence promoting the performance in the high-level tasks.

\section{Conclusion}\label{conclusion}

The ever-growing number of multimodal RS data poses not only a challenge but also an opportunity to the EO tasks. By jointly utilizing their complementary features, great breakthroughs have been witnessed over the recent years. Particularly, artificial intelligence-related technologies has demonstrated their advantages over traditional methods on account of their superiority in feature extraction. Driven by aforementioned RS big data and cutting-edge tools, DL-based multimodal RS data fusion becomes a significant topic in the RS community. Therefore, this review gives a comprehensive introduction on this fast-growing domain, including a literature analysis, a systematic summary in several prevalent sub-fields in RS fusion, a list of available resources, and the prospects for the future development. Specifically, we focus on the second part, i.e., DL-based methods in different fusion subdomains, and give a detailed study in terms of used models, tasks, and data types. Finally, we are encouraging to find that DL has been applied to every corner of multimodal RS data fusion and obtains tremendous and promising achievements in recent years, which provide researchers more confidences to conduct in-depth study in the future.

\section*{Acknowledgements}  

This work was supported by the National Natural Science Foundation of China [62161160336, 42030111]; MIAI@Grenoble Alpes [ANR-19-P3IA-0003]; and the AXA Research Fund.

\bibliography{reference}

\end{document}